\pdfoutput=1
\documentclass[11pt]{article}

\usepackage{acl}
\usepackage[inkscapelatex=false]{svg}

\usepackage{adjustbox}
\usepackage{times}
\usepackage{latexsym}
\usepackage{amssymb}
\usepackage{multirow}
\usepackage{multicol}
\usepackage{booktabs}
\usepackage{soul}
\usepackage{xcolor}
\usepackage{colortbl}
\usepackage{listings}
\usepackage{courier}
\usepackage{url}
\usepackage{sourcecodepro}
\usepackage[most]{tcolorbox}
\usepackage{listings}
\usepackage{fancybox}
\usepackage{enumitem}
\usepackage{pifont}
\usepackage{caption}
\usepackage{subcaption}
\usepackage{hyperref}
\usepackage[T1]{fontenc}
\usepackage[utf8]{inputenc}

\usepackage{microtype}

\title{DocFinQA: A Long-Context Financial Reasoning Dataset}

\author{Varshini Reddy, Rik Koncel-Kedziorski, Viet Dac Lai\\
\textbf{ Michael Krumdick, Charles Lovering, Chris Tanner} \\
  Kensho Technologies\\
  \texttt{varshini.bogolu@kensho.com}
  }

\begin{document}
\maketitle
\begin{abstract}
For large language models (LLMs) to be effective in the financial domain -- where each decision can have a significant impact -- it is necessary to investigate realistic tasks and data.
Financial professionals often interact with documents spanning hundreds of pages, but most financial research datasets only deal with short excerpts from these documents. 
To address this, we introduce a long-document financial QA task. 
We augment 7,437 questions from the existing FinQA dataset with full-document context, extending the average context length from under 700 words in FinQA to 123k words in DocFinQA. 
We conduct extensive experiments over retrieval-based QA pipelines and long-context language models. Based on our experiments, DocFinQA proves a significant challenge for even state-of-the-art systems. 
We also provide a case study on a subset of the longest documents in DocFinQA and find that models particularly struggle with these documents. 
Addressing these challenges may have a wide-reaching impact across applications where specificity and long-range contexts are critical, like gene sequences and legal document contract analysis. 
DocFinQA dataset is publicly accessible\footnote{ \scriptsize \url{https://huggingface.co/datasets/kensho/DocFinQA}}.

\end{abstract}

\newcommand{\mtr}[2]{\multirow{#1}{*}{\textbf{#2}}}
\newcommand{\mtc}[2]{\multicolumn{#1}{c}{\textbf{#2}}}
\newcommand{\mtcb}[2]{\multicolumn{#1}{|c|}{\textbf{#2}}} % with border
\newcommand{\mrt}[2]{\multirow{#1}{*}{\rotatebox{90}{#2}}}

\definecolor{C_SECOND_SOFT}{HTML}{e2a8d6}
\definecolor{C_BASE_SOFT}{HTML}{66adbf}

\definecolor{C_BASE}{HTML}{a2c4c9}
\definecolor{C_SECOND}{HTML}{B62699}
\definecolor{C_THIRD}{HTML}{00B9E8}
\definecolor{C_FOURTH}{HTML}{0B008F}
\definecolor{C_BACKGROUND}{HTML}{EDF5FC}

\section{Introduction}
\label{sec:introduction}

The frequent need to reason over large volumes of textual and tabular data makes financial analysis particularly challenging for LLMs~\cite{azzi-etal-2019-finsbd}.
Existing work on automating financial numerical reasoning focuses on unrealistically specific document snippets \cite{chen-etal-2021-finqa,zhu-etal-2021-tat}. Datasets are often limited to pre-selected document sections, failing to reflect the broader and more realistic scenarios faced by analysts \cite{masson-montariol-2020-detecting}. Financial professionals usually sift through hundreds of pages per document, requiring a deep understanding of both content and structure to navigate and extract pertinent information effectively. Current long-document QA datasets such as NarrativeQA \citet{kocisky-etal-2018-narrativeqa} do not test the quantitative reasoning skills needed in the financial domain.

\begin{figure}[t]
    \centering
    \includegraphics[width=\linewidth]{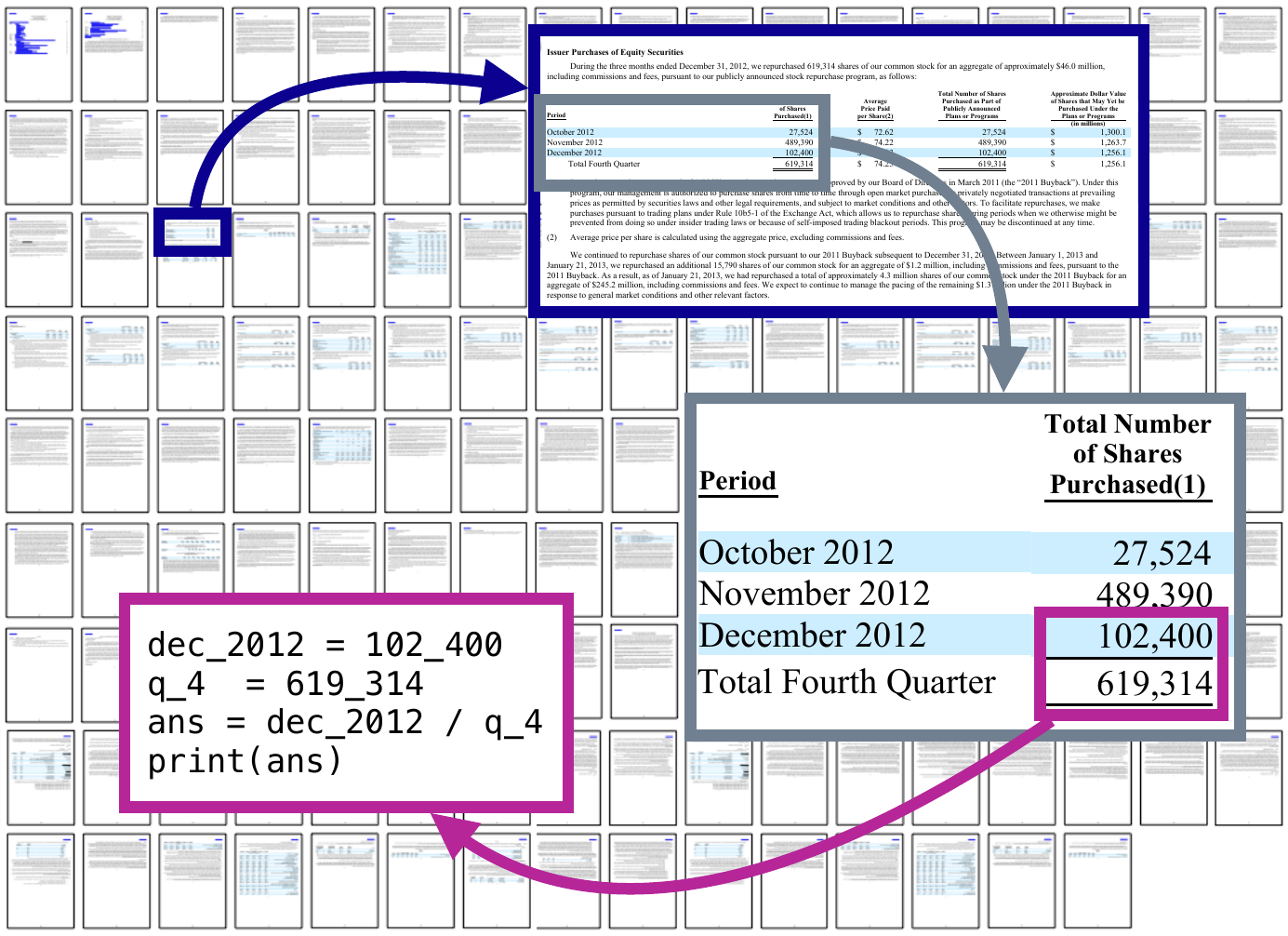}
    \caption{%DocFinQA extends FinQA to inputs with over 100K tokens. 
    DocFinQA extends FinQA to documents often over 150 pages long (100K+ tokens), so it is difficult to find the pertinent information. The question for the example above is: ``For the quarter December 31, 2012 what was the percent of the total number of shares purchased in December?'' The correct answer is 16.5\%.}
    \label{fig:teaser}
\end{figure}

In this work, we introduce DocFinQA, a long-document financial question-answering task. 
We extend the FinQA dataset of expert annotated questions and answers \cite{chen-etal-2021-finqa} with full Securities and Exchange Commission (SEC) reports. This results in a significantly longer context in the DocFinQA dataset -- by a factor of 175 -- than the FinQA dataset. Additionally, we manually verified and annotated questions of the test set. The resulting long-document QA task offers a more realistic evaluation of a model's reasoning capabilities over financial documents.
In line with recent work on program synthesis for financial QA \cite{bizbench}, the questions in DocFinQA are appended with Python programs to generate the answers, allowing for training and evaluating program synthesis models for use in realistic financial workflows.

Using this setup, we evaluate retrieval-based and long-context LLM systems. We study a typical retrieval pipeline that chunks and encodes the document, searching for the best chunks given a question, and passing the question and top-$k$ chunks to a generative QA model \cite{hsu-etal-2021-answer}.

We also evaluate retrieval-free approaches using long-context LLMs \cite{weston2023system}.
Our results show that the successful employment of LLMs in financial settings requires further study of the specific nuances of the financial domain, such as context disambiguation. Our dataset represents a step towards better capturing these nuances.

\section{Related Work}
\label{sec:related-work}

Prior studies in financial question answering focus on non-numerical reasoning \cite{day2016deep,jorgensen-etal-2023-multifin,maia201818}. Short-context grounded numerical reasoning tasks were introduced with datasets such as FinQA \cite{chen-etal-2021-finqa} and TAT-QA \cite{zhu-etal-2021-tat}. Recently, understanding long documents has attracted more attention for tasks involving events \cite{yang-etal-2018-dcfee}, table of contents \cite{bentabet-etal-2020-financial}, and causal relations \cite{mariko-etal-2022-financial}. However, to the best of our knowledge, this is the first attempt to address financial numerical QA grounded in long documents with upwards of hundreds of pages of context for each question.

Comparison of DocFinQA and existing Finance QA and Long Document QA Long-document QA has been studied in NLP with the introduction of datasets such as SearchQA \cite{dunn2017searchqa}, NarrativeQA \cite{kocisky-etal-2018-narrativeqa}, QuALITY \cite{pang-etal-2022-quality}, and PDFTriage \cite{saad2023pdftriage}. Due to the limited context size of LLMs, retrieval-based models are commonly used to filter irrelevant text \cite{izacard2022few,lewis2020retrieval}. Recently, advances in attention mechanisms \cite{beltagy2020longformer,dao2022flashattention} and positional embeddings \cite{press2021train,su2023roformer} allow for end-to-end grounded QA with context windows of more than 100k tokens. However, these methods suffer from loss of important context \cite{zhang-etal-2023-extractive} and often fail to make full use of longer inputs \cite{liu2023lost}. Our work studies the intersection of numerical reasoning and long-document processing, and our results demonstrate that there is still ample room for improvement in this domain.

\section{DocFinQA Dataset}

% FinQA was created by finance experts from S\&P 500 companies' annual financial reports (aka., 10-K). They used full documents when creating the questions (e.g., 100s of pages), but only included the curated context necessary to produce correct answers (e.g., a paragraph and/or table of statistics). An example of FinQA is shown in Table~\ref{tab:appendix_example_finqa}.

% To recreate the original problem, we retrieve the full SEC filing for each question. Following \cite{bizbench}, we require the model to generate Python code to produce an answer instead of directly generating the numeric answer. The generated Python code illustrates what information from the context is used, along with what arithmetic operations are performed. 

{\bf Dataset Representation} Each question in FinQA is a triplet $(c^{golden},q,a)$ composed of a golden context $c^{golden}$, a  question $q$, and an answer $a$ written in human language. An example of FinQA is shown in Table~\ref{tab:appendix_example_finqa} (See Appendix \ref{app:sec-filings}). We extend the dataset in two ways: (1) context $c^{golden}$ is extended to the full document context $D$, and (2) we added a Python program $p$ that produces the answer $a$.
Each final sample in DocFinQA is a quartet $(D, q, p, a)$. An example of DocFinQA is shown in Table~\ref{tab:appendix_example_docfinqa} (See Appendix \ref{app:sec-filings}).

\begin{figure}[!h]
    \centering
    \includegraphics[width=\linewidth]{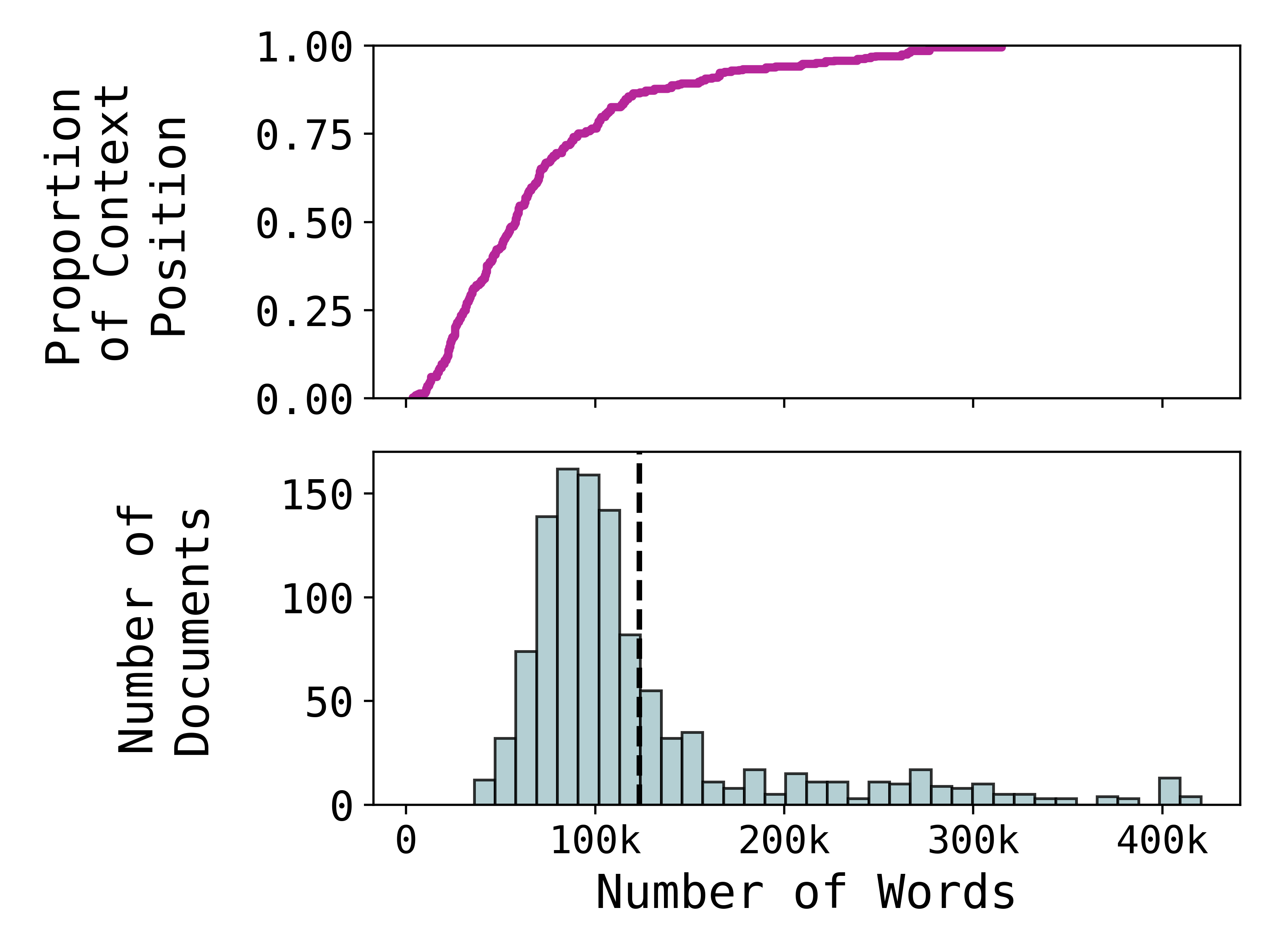}
    \caption{Histogram of document length (\#words) in DocFinQA dataset with dash line representing the average length of the documents. The purple line depicts the proportion of documents where the question context is within the current number of words.}
    \label{fig:histogram}
\end{figure}

{\bf Filings Collection:} For each question of the FinQA dataset, we identify the corresponding SEC filing from which it was created. We retrieve the filing in HTML/XML format from \href{https://www.sec.gov/edgar/}{SEC's EDGAR service} and parse the text and table into clean markdown format \cite{wang2023graphical}. The collection and parsing processes are presented in more detail in Appendix~\ref{app:sec-filings} and Appendix~\ref{app:sec-parsing}, respectively. Figure \ref{fig:histogram} shows the distribution of document lengths in DocFinQA.
\\

% \definecolor{orange}{HTML}{FF7F00}

\newcommand{\yes}[0]{$\color{violet}\checkmark$}
\newcommand{\nah}[0]{$\color{red}$}

% \begin{table}[t]
% \resizebox{\linewidth}{!}{
%     \begin{tabular}{l|rrr|ccc}
%         \toprule
%         {\bf Dataset}      & {\bf \#Docs} & {\bf \#QAs}  & {\bf \#Words} & {\bf Page} & {\bf Number} & {\bf Table} \\
%         \midrule
%         NarrativeQA & 1,572 & 46,765   & 63,000  & \yes  &   -   &   -   \\
%         QuALITY     &  381  & 6,737    & 5,159  & \yes  &   -   &   -   \\
%         PDFTriage   & 82    & 908     & 12,000    & \yes  &  \yes & \yes  \\
%         \midrule
%         TAT-QA      & 2,757 & 16,552    & 260 & -     & \yes  & \yes\\
%         FinQA       & 2,789 & 8,281    & 687  & -     & \yes  & \yes\\
%         \midrule
%         DocFinQA    & 801 & 7,437  & 123,453   & \yes  & \yes  & \yes \\
%         \bottomrule
%     \end{tabular}
% }
% \caption{Comparison of DocFinQA and existing Finance QA and Long Document QA dataset. DocFinQA includes multi-{\bf page} documents rich with both {\bf numeric} and {\bf table} data.}
% \label{tab:dataset}
% \end{table}

\begin{table*}[t]
    \small \centering
    \begin{tabular}{l|rrr|ccc}
        \toprule
        {\bf Dataset}      & {\bf \#Docs} & {\bf \#QAs}  & {\bf \#Words} & {\bf Multi-page} & {\bf Numeric} & {\bf Tabular} \\
        \midrule
        NarrativeQA & 1,572 & 46,765   & 63,000  & \yes  &   -   &   -   \\
        QuALITY     &  381  & 6,737    & 5,159  & \yes  &   -   &   -   \\
        PDFTriage   & 82    & 908     & 12,000    & \yes  &  \yes & \yes  \\
        \midrule
        TAT-QA      & 2,757 & 16,552    & 260 & -     & \yes  & \yes\\
        FinQA       & 2,789 & 8,281    & 687  & -     & \yes  & \yes\\
        \midrule
        DocFinQA    & 801 & 7,437  & 123,453   & \yes  & \yes  & \yes \\
        \bottomrule
    \end{tabular}
\caption{Comparison of DocFinQA and existing Finance QA and Long Document QA dataset. DocFinQA includes {\bf multi-page} documents with both {\bf numeric} and {\bf tabular} data.}
\label{tab:dataset}
\end{table*}

\textbf{Chunking and Alignment:}  
To study retrieval-based QA systems, we split each document $D$ into a set of chunks $C=\{c_1,\cdots,c_n\}$. Each chunk consists of 2,750 characters ($\sim\!509$ tokens) with a 20\% overlap to avoid missing context at the edges. To compute the performance, we identify the best context chunk, $c^{\bigstar}$, from the chunk set $C$ associated with each document $D$ that includes the information to answer question $q$. Since FinQA already provides $c^{golden}$, we compute a pair-wise score $(c_i, c^{golden})$, for all chunks, $c_i \in C$, including the golden chunk. We find that four-gram-based similarity score offers the sharpest matching signal among tri-gram, four-gram, and fuzzy matching. The chunk with the highest score is selected as the target context chunk for retrieval. We verify that this process results in good $c^{\bigstar}$ chunks through manual inspection and by substituting $c^{\bigstar}$ for $c^{golden}$ in a few-shot QA evaluation with GPT-3.5. 
\\

\textbf{Code Generation:} 
\label{sec:code-generation}
The FinQA dataset provides solutions in a ``program'' syntax that, when executed, yields the answer (e.g., in Figure \ref{fig:teaser} the solution is \texttt{divide(102400, 619314)}. However, this derivation does not provide meaningful context of what is being calculated. In our running example, 102400 is not semantically grounded to the document.  \citet{bizbench} augments FinQA with readable Python code (including named variables like, \texttt{dec\_shares = 102\_400}) that can be executed to derive the answer, providing a layer of interpretability. Thus, we use the code-enhanced version of  DocFinQA (See Appendix \ref{app:code-generation}).
\\

{\bf Statistics:} The resultant DocFinQA dataset comprises of 5,735 training, 780 development, and 922 test samples, derived from 801 unique SEC filings. Table \ref{tab:dataset} shows the statistics and characteristics of DocFinQA in comparison with other finance numerical reasoning and long-document QA datasets.
\\

\textbf{Impact of Data Selection - DocFinQA vs FinQA:} Due to the limited availability of complete SEC filings (refer Appendix \ref{app:sec-filings}) and imperfections in the code generation process, DocFinQA encompasses 7,437 out of 8,281 of FinQA questions. This process may filter out a collection of question types that the LLM did not answer due to its limited capability. We investigates the impact of this process by comparing the distribution of the question types in FinQA and DocFinQA. To do this, we show the distribution of questions grouped by their first 2 non-stop words in Figure \ref{fig:question-dist} (Appendix \ref{app:questio-dist}). The most important observation is that, overall, the distribution of the question set in DocFinQA and FinQA are very similar. No major groups are being filtered out by our data selection process. The dominant questions (above 1\% in FinQA) remain dominant and no major impact on the percentages of those questions is observed. The mid-group (above 0.2\% in FinQA) question sets see a mixed effect. A large portion of these questions are increased in percentage while some experience significant loss (e.g., ``{\it what percentual}'' and ``{\it what decrease}''). Lastly, the long tail group ( under 0.2\% in FinQA) either remains the same (e.g., ``{percent total}'' and ``{\it what greatest}'') or is completely wiped out due to a small population (e.g., ``{\it was average}'', and ``{\it what return}'').

\begin{table*}[]
\resizebox{\textwidth}{!}{
\begin{tabular}{lrrrrrrrrrrrrrrr}%{lc|cc|ccc|ccc|ccc|ccc}
    \hline
    \mtr{3.5}{Model} & \mtr{3.5}{\bf Size} & \mtc{2}{Upper Bound}  & \mtc{3}{Original ColBERT} & \mtc{3}{Finetuned ColBERT} & \mtc{3}{Sentence-BERT} & \mtc{3}{OpenAI ADA}\\
    \cmidrule(lr){3-4}
    \cmidrule(lr){5-7}
    \cmidrule(lr){8-10}
    \cmidrule(lr){11-13}
    \cmidrule(lr){14-16}
    % \cmidrule(lr){17-19}
    & & \mtc{1}{*} & \mtc{1}{*} & \mtc{1}{Top 1} & \mtc{1}{Top 3} & \mtc{1}{Top 3} & \mtc{1}{Top 1} & \mtc{1}{Top 3} & \mtc{1}{Top 3} & \mtc{1}{Top 1} & \mtc{1}{Top 3} & \mtc{1}{Top 3} & \mtc{1}{Top 1} & \mtc{1}{Top 3} & \mtc{1}{Top 3} \\
    & & \mtc{1}{1 shot} & \mtc{1}{3 shot} & \mtc{1}{3 shot} & \mtc{1}{1 shot} & \mtc{1}{3 shot}  & \mtc{1}{3 shot}  & \mtc{1}{1 shot} & \mtc{1}{3 shot} &  \mtc{1}{3 shot}  & \mtc{1}{1 shot} & \mtc{1}{3 shot} &   \mtc{1}{3 shot}  & \mtc{1}{1 shot} & \mtc{1}{3 shot}  \\
    % \hline
    % RANDOM         &         & 0.0  & 0.0  & \multicolumn{1}{c}{0.0} & \multicolumn{1}{c}{0.0}  & 0.0  & 0.0  \\
    \hline
    Falcon           & 7B    & 2.0 & 2.0 & \underline{\color{violet}1.9} & 0.0 & 0.0 & \underline{\color{violet}1.9} & 1.3 & 0.0 & \underline{\color{violet}1.2} & 0.1 & 1.3 & \color{violet}\textbf{2.0} & 0.1 & 0.0  \\
    \hline
    MPT              & 7B     & 6.8 & 6.6 & \underline{\color{violet}4.5} & 0.8 & 0.2 & \color{violet}\textbf{4.9} & 1.0 & 1.2 & \underline{\color{violet}3.9} & 0.6 & 0.8 & \underline{\color{violet}4.3} & 1.6 & 2.0 \\
    MPT              & 30B   & 27.1 & 31.0 & \underline{\color{violet}15.3} & 2.2 & 1.7 & \color{violet}\textbf{16.8} & 3.2 & \underline{\color{violet}3.8} & 1.1 & 3.8 & 2.7 & \underline{\color{violet}15.7} & 10.4 & 5.1 \\
    % \hline
    % StarCoder        & 16   & 26.3 & 31.2 & 13.0 & 13.8 & 16.0 & 14.7 & 16.8 & \textbf{17.5} & 10.4 & 9.1 & 11.6 & 14.1 & 12.1 & 16.1 \\
    \hline
    Llama 2          & 7B    & 17.3 & 22.0 & \underline{\color{violet}12.8} & 5.8 & 8.0 & \color{violet}\textbf{14.0} & 6.0 & 10.3 & \underline{\color{violet}8.9} & 2.7 & 6.5 & \underline{\color{violet}11.2} & 4.0 & 11.0 \\
    Llama 2 + SFT & 7B   & 67.1 & 69.7 & 30.0 & \underline{\color{violet}32.6} & 31.3 & 32.2 & \color{violet}\textbf{35.3} & 33.9 & 19.9 & 24.1 & \underline{\color{violet}24.3} & 28.7 & \underline{\color{violet}29.4} & 27.7  \\

    Llama 2          & 13B   & 30.0 & 33.4 & \underline{\color{violet}14.4} & 10.4 & 14.1 & \color{violet}\textbf{19.1} & 11.9 & 14.5 & \underline{\color{violet}14.9} & 7.9 & 10.2 & \underline{\color{violet}18.3} & 9.8 & 13.7  \\

    \hline
    CodeLlama        & 7B    & 26.9 & 34.0 & 12.6 & 11.4 & \underline{\color{violet}16.1} & 15.7 & 12.3 & \underline{\color{violet}16.8} & 11.9 & 8.9 & \underline{\color{violet}13.2} & 15.4 & 14.2 & \color{violet}\textbf{17.5} \\
    CodeLlama        & 13B   & 32.1 & 39.0 & 19.5 & 14.8 & \underline{\color{violet}21.5} & 21.2 & 15.7 & \color{violet}\textbf{22.5} & 13.2 & 8.5 & \underline{\color{violet}16.0} & 18.3 & 14.4 & \underline{\color{violet}20.9} \\
    \hline
    
    Mistral          & 7B    & 39.7 & 48.8 & \underline{\color{violet}23.0} & 18.8 & 21.3 & \color{violet}\textbf{25.9} & 16.8 & 25.2 & \underline{\color{violet}19.0} & 13.6 & 17.6 & 20.9 & 18.8 & \underline{\color{violet}22.1} \\
    \hline
    % GPT 3            & 175B  & 58.4 & 61.5 & \color{violet}30.6 & 6.5 & 6.5 & \color{violet}\textbf{32.9} & 8.7 & 9.4 & \color{violet}21.6 & 1.0 & 7.8 & \color{violet}28.6 & 16.2 & 15.3 \\
    GPT 3.5          & -  & 67.3 & 67.5 & 36.0 & \underline{\color{violet}39.0} & 38.8 & 38.8 & \color{violet}\textbf{40.7} & 40.2 & 24.8 & 30.1 & \underline{\color{violet}36.3} & 35.0 & 36.5 & \underline{\color{violet}36.9} \\

    \hline
\end{tabular}
}
\caption{Performance of the models on DocFinQA in one-shot and few-shot in-context learning settings for the top 1 and top 3 retrieved chunk contexts on the development set. For each model, the best performance among all configurations is in \textbf{\textcolor{violet}{bold}}. For each model, the best performance among different configurations for the same retrieval model is \underline{\textcolor{violet}{underlined}}. Top 1 and Top 3 indicate the number of retrieved chunks used as context for a configuration. *The single original context chunk from the FinQA test set is used to estimate the upper bound.}
\label{tab:downstream-dev}
\end{table*}

\section{Retrieval-based QA Evaluation}

\label{sec:retrieval-based}

\quad {\bf Retrieval Task:} We test three models for context retrieval: ColBERT ({\bf ColB})\cite{khattab2020colbert}, Sentence-BERT ({\bf SentB}) \cite{reimers-gurevych-2019-sentence}, and OpenAI's {\bf Ada} \cite{openai-ada}. Further, we finetune the ColBERT model ({\bf FT ColB}) on the training set of DocFinQA to evaluate an in-domain model. More details on the fine-tuning process are given in Appendix \ref{app:colber-finetuning}. We also test a matching-based model, BM25 \cite{robertson1995okapi}, but observe poor performance (See  Appendix \ref{app:bm25-included} for details). 

\begin{figure}[!h]
 \centering
    \includegraphics[width=\linewidth]{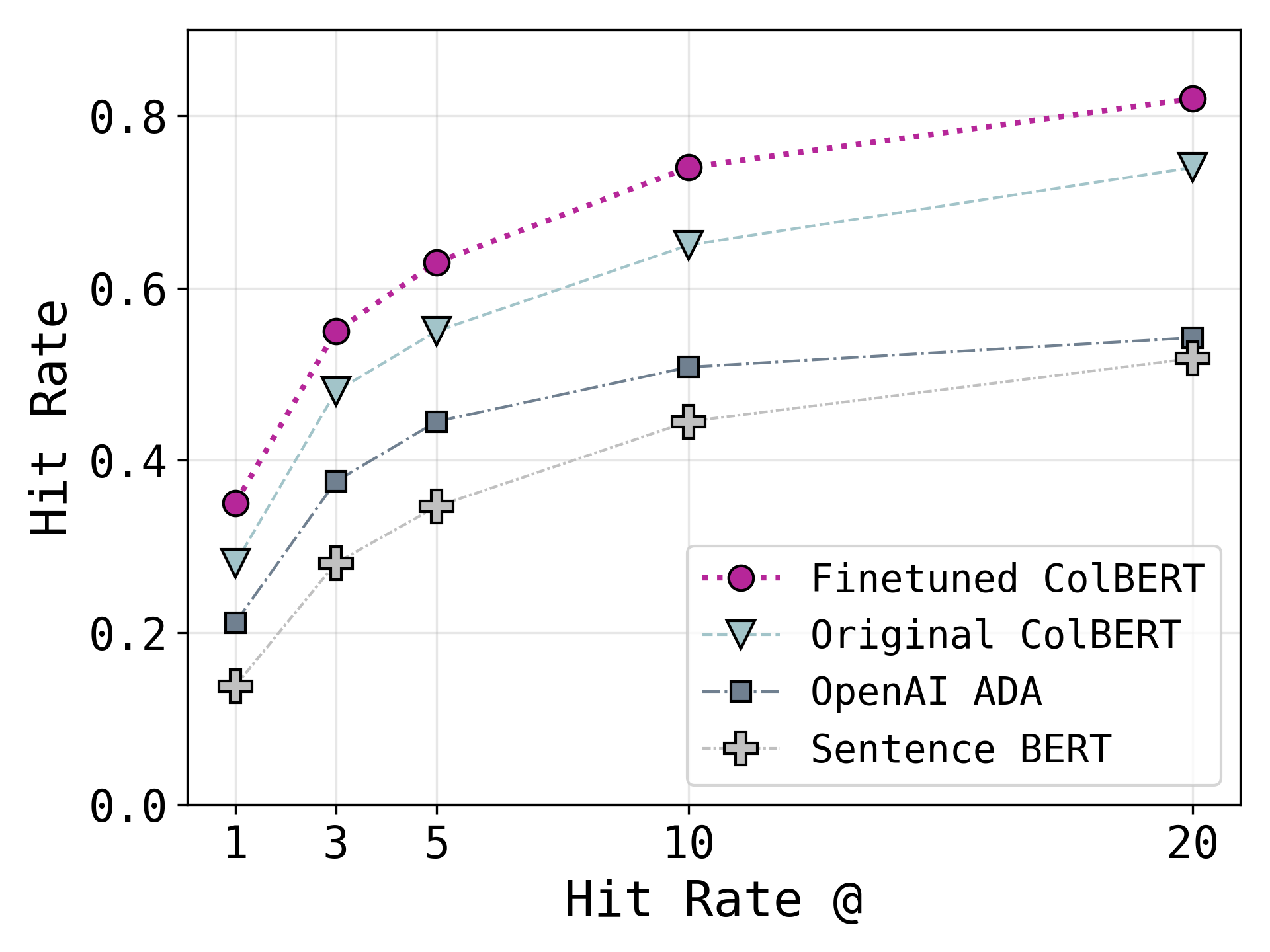}
    \caption{Hit rate of retrieval models. 
    }
    \label{fig:hit_rate_cosine}
\end{figure}

To retrieve context for a question $q$ over chunk set $C$, we encode both $q$ and $C$ with the encoding models mentioned above. This results in an embedding, $v_q$, for the question and chunk embeddings $V_C=\{v_{c_i}|c_i\in C\}$.
We compute the cosine similarity between  $v_q$ and each vector in $V_C$ to retrieve the top-k most similar chunks. We evaluate these models using HR@k on the test set of DocFinQA using the target $c^{\bigstar}$. Results are shown in Figure~\ref{fig:hit_rate_cosine}. The {\bf FT ColB}  yields the highest HR, followed by {\bf ColB}. {\bf FT ColB} yields an average improvement of 91\% HR over {\bf SentB} and obtains a 0.35 (HR@1) and 0.55 (HR@3). 
\\

\subsection{Question Answering Task} We formulate the QA task as a few-shot in-context learning task \cite{brown2020language}. For each in-context example, we only provide the relevant chunk and the answer. For the actual query, we provide $k$ chunks. More details of the few-shot settings are provided in Appendix~\ref{app:few-shot}.

We evaluate Falcon \cite{penedo2023refinedweb}, MPT \cite{mosaicml2023mpt}, LLaMa 2 and CodeLlaMa \cite{touvron2023llama}, Mistral \cite{jiang2023mistral}, GPT3.5 \cite{brown2020language,openai2023gpt4} models. We weren't able to evaluate proprietary models such as GPT3 \cite{brown2020language}, BloombergGPT \cite{wu2023bloomberggpt} due to their inaccessibility. We skipped models that were not finetuned for code generation such as PIXIU \cite{xie2023pixiu} and FinGPT \cite{yang2023fingpt} due to their poor performance. We also skipped models trained for other languages such as BBT-Fin \cite{lu2023bbt} and XuanYuan 2.0 \cite{zhang2023xuanyuan}.
\\ 

\textbf{Performance on the development set}: Table \ref{tab:downstream-dev} reports the full performance of the development set with four retrieval models and three few-shot settings. This results in a total of twelve unique configurations. For both fine-tuned and pre-trained models, we use greedy decoding whenever applicable. One trend noted was that all generic LLMs showed higher accuracy with shorter context and more few-shot examples i.e. top chunk with 3 shots. While code-based LLMs such as Starcoder and CodeLLama showed higher accuracy with longer context i.e. top 3 chunks with 3 shots. This trend is also depicted in Figure~\ref{fig:topk-numshot}.
\\ 

\textbf{Performance on the test set}: Table \ref{tab:result-retrieval-based} reports the performance of the same 10 state-of-the-art models on the test set of DocFinQA. The few-shot setting and retrieval model configuration for each LLM are treated as hyperparameters and are picked based on the performance of the development set. 
We observe that larger models outperform smaller models (e.g., MPT 30B vs MPT 7B). Models trained on code yield higher accuracy than non-code models (e.g., CodeLlama vs Llama). Models with additional supervised finetuning (e.g., LLama 2/7B+SFT) and instruction tuning (e.g., GPT-3.5) are among the best examined. Notably, Mistral 7B outperforms several larger models, although it lags behind Llama 2/7B+SFT and GPT-3.5.

The {\bf FT ColB} model is the best retrieval model in all but one setting. It yields a marginal but consistent improvement over the {\bf ColB}, and a large improvement over {\bf SentB} and {\bf Ada}.

\begin{table}[]
\centering
\begingroup
\small	
\setlength{\tabcolsep}{3pt} 
\begin{tabular}{l|ccc|c}
    \toprule
    \mtr{1}{Model/Size} & \bf ColB  & \bf SentB & \bf ADA & \bf \textbf{FT ColB} \\
    % & \bf ColB. &\bf  B dERT &\bf  ADA & \bf  ColB. \\
    \midrule
    % GPT 3.5 (FinQA) & \multicolumn{4}{c}{67.5} \\
    % \midrule
    Falcon/7B       & \;\;\textbf{2.3} &  \;\;0.3 & \;\;1.2  & \;\;1.8\\
    \midrule
    MPT/7B          & \;\;4.6          & \;\;2.7  & \;\;3.8  & \;\;\textbf{4.8}\\
    MPT/30B         & 17.3         & 11.1 & 12.1 & \textbf{18.1}\\
    \midrule
    Llama 2/7B      & 13.5         & 8.9  & 11.1 & \textbf{13.5}\\
    Llama 2/13B     & 18.7         & 12.7 & 14.9 & \textbf{19.1}\\
    \midrule
    CodeLlama/7B    & 15.6         & 12.2 & 15.2 & \textbf{16.8}\\
    CodeLlama/13B   & 19.1         & 13.8 & 18.8 & \textbf{21.0} \\
    \midrule
    Mistral/7B      & 23.2         & 14.9 & 21.5 & \textbf{25.0}\\
    \midrule
    Llama 2/7B+SFT  & 32.9         & 24.8 & 34.3 & \textbf{36.1}\\
    \midrule
    % GPT 3 /175B     & 32.33         & 23.16 & 27.58 & \textbf{32.79}\\
    GPT-3.5/-       & 41.6 & 33.8 & 36.4 & \textbf{42.6}\\
    \bottomrule
\end{tabular}
\endgroup
\caption{Performance on DocFinQA test set.
% with the best hyperparameter settings obtained from development set experiments (Refer Table~\ref{tab:downstream-dev}). 
For each row, the best performance among all retrieval models is
in \textbf{bold}. The fewshot setting is selected based on the best performance on the development set (See Table \ref{tab:downstream-dev}).
% The performance of GPT 3.5 on FinQA is added for comparison. 
}
\label{tab:result-retrieval-based}
\end{table}

% \section{Retrieval-free QA Evaluation}
\section{Case Study w/ 100K+ Token Documents}
\label{sec:retfree}

% * hardest case
% * 40\% require more than 100K tokens
% * we can note for a similar sample of $<100k$ documents performance is indeed on par with humans at 46.5

Recent LLMs can handle context lengths of 128K tokens, but more than 40\% of the documents in DocFinQA remain unanswerable even at this content length (see Figure \ref{fig:histogram}). Here, we evaluate performance on a \textbf{test subsample} of 200 randomly selected documents, each of which has 100K or more tokens due to the monetary and temporal costs of human evaluation and GPT4.

We explore two retrieval-free options - System 2 Attention ({\bf S2A}) and {\bf Iterative} method. S2A extracts relevant information from each 100K-token chunk of a document before answering the question using the combined extracted information as context ~\citep{weston2023system}. The Iterative method produces the output program iteratively as the LLM processes each 100k section of the document. A temporary answer program (initially ``None'') is input with each section to the LLM. We also report the performance of the best retrieval-based model ({\bf Retrieval}) based on the experiment in Section \ref{sec:retrieval-based}. 

We conducted human evaluations on these 200 questions highlighting the challenging nature of this dataset with experienced but non-expert human participants (See Appendix~\ref{app:human-eval} for details). Non-expert human performance on DocFinQA is lower than human performance reported in FinQA \cite{chen-etal-2021-finqa} (41\% versus 50.7\%). This can be attributed to the difficulty of finding the golden page, compared to the golden page being given in FinQA. Notably, the expert performance reported in FinQA is 91.2\%.

\begin{table}[]
\centering
\begingroup
\small
\setlength{\tabcolsep}{3pt} 
\begin{tabular}{l|c|c}
    
    \toprule
    \bf Model/Size + Method  & \bf w/ Retrieval & \bf Test Subsample  \\
     % &  & \bf Interactive & \bf S2A &\\
    \midrule
    Human  & No & 41.0 \\ % 43.9 (over 187) 41(over 200)
    \midrule
    Mistral/7B + Iterative  & No   &  11.5 \\ % 13.1
    Mistral/7B + S2A         & No   &  15.5 \\
    \midrule
    Mistral/7B + Retrieval   & Yes   &  20.0 \\
    \midrule
    GPT-4 + Iterative       & No   & 20.0 \\
    GPT-4 + S2A              & No   & 23.0 \\
    \midrule
    GPT-4 + Retrieval        & Yes   & 47.5 \\
    \bottomrule
\end{tabular}
\endgroup
\caption{Retrieval-free performance on a case-study of $100K+$ token documents. 
% Humans  surpass retrieval-free LLMs.
% on full context documents
% The performance GPT-4 on FinQA and other retrieval-based models are added for comparison.
% with the best hyperparameter settings obtained from development set experiments (Refer Table \ref{tab:downstream-dev}). 
%The performance of GPT-4 on FinQA is added for comparison. 
} 
\label{tab:ressult-retrieval-free}
\end{table}

% \begin{table}[]
% \centering
% \begingroup
% \small
% \setlength{\tabcolsep}{3pt} 
% \begin{tabular}{l|c|c}
%     \toprule
%     \textbf{Model}  & \bf Test Subsample %\textcolor{red}{(todo: long)} &  \textbf{Avg}\\
%     \midrule
%     Mistral/7B  & 20.6 \\ % & 20.3 \\
%     % LLama 2/7B+SFT   & 0.5 &  30.1\\
%     % GPT-3.5 & 36.2 & 42.3\\
%     GPT-4 & 47.5 \\ % & \bf 52.2\\
%     \bottomrule
% \end{tabular}
% \endgroup
% \caption{Performance on DocFinQA test subsample improves with retrieval. 
% %where human performance on full context documents surpasses all state of the art LLMs.
% % The performance GPT-4 on FinQA and other retrieval-based models are added for comparison.
% % with the best hyperparameter settings obtained from development set experiments (Refer Table \ref{tab:downstream-dev}). 
% %The performance of GPT-4 on FinQA is added for comparison. 
% } 
% \label{tab:ressult-retrieval-with}
% \end{table}

Nonetheless, the non-expert human performance is double that of retrieval-free GPT-4 on these long documents, and roughly triple that of retrieval-free Mistral models. The performance of the iterative method was worse than S2A for both GPT-4 and Mistral with a reduced accuracy of 3\% and 4\%, respectively. With retrieval, both Mistral and GPT-4 outperform their retrieval-free counterparts, with the assisted GPT-4 now on par with the human cohort.
% These results for different long document combination methods suggests that the field needs to further explore methods for combining information across multiple calls to a document-processing LM.
Together, these results highlight that DocFinQA is a difficult test for long-document QA and that there is still room for significant improvement in this domain. For instance, further exploration into methods that combine information across multiple calls to a document-processing LLM is warranted.

\section{Conclusion}

This paper introduces a realistic document-level question-answering dataset over financial reports. Each question includes a full financial report (averaging 123K words), a far greater challenge than previous work that hones in on pre-specified content. Our findings reveal that this more realistic setting presents a significantly more difficult challenge, thereby opening new avenues for research in quantitative financial question answering.

\clearpage

\section*{Acknowledgement}

We express our gratitude to our colleagues at Kensho Technologies and S\&P Global for their invaluable contributions to data annotation, which greatly enhanced the completion of this project. We extend our appreciation to the anonymous reviewers for their supportive input.

\section*{Limitation}
\label{sec:limitation}
This work introduced an extension of the existing FinQA dataset. Due to limited human resources, we only validated the test set while the training and the development set were not fully validated. As a result, we can not make any claim of bias and question quality in the not-yet-validated data points offered in this paper. Additionally, as discussed in section \ref{sec:code-generation}, the code provided in this work was generated by WizardCoder LLMs. We assume that the code is correct if it produces correct or approximately close to the golden answer. This method may generate both false positive codes (the code that generates the correct answer with incorrect rationales) and false negative codes (the correct code that fails the approximation test).

\section*{Broader Impact and Ethical Considerations}
\label{sec:ethical-condisderation}
We do not foresee any considerable risks associated with our work given that it is an extension of an open-source dataset and uses publicly available documents. To uphold transparency, the paper provides detailed documentation of the dataset creation process, including the sources of data and annotation details. Our dataset serves as a resource to underscore the need for longer context-oriented benchmarks both within and outside the financial domain and does not intend to criticize one or more LLMs. 
% Additionally, as this work evaluates existing models on our dataset, we are not aware of any potential negative impact.

The annotation in this work is done automatically or in-house, so no crowd-sourced or contract annotators were hired throughout the process.
The human evaluation in this study was done by full-time paid coworkers known to the authors.

\bibliography{anthology,custom}
\bibliographystyle{acl_natbib}

\appendix

\clearpage

\section{SEC Filing Collection}
\label{app:sec-filings}
Each data point in the FinQA dataset consists of a document identification field as shown in Table~\ref{tab:appendix_example_finqa}. This field is made up of 3 sections separated by a forward slash. The first is a string called company ticker symbol, the second refers to the year in which this document was filed and the third is the page number in the document where the answer can be found. 

Downloading the right 10-K filing from the SEC begins with identifying the company code from the company ticker symbol. For example, \texttt{C/2017/page\_328.pdf-1} in FinQA maps to the \texttt{CITIGROUP INC} with company code \texttt{831001}. This mapping is obtained from the official file released by SEC which can be found here \url{https://www.sec.gov/file/company-tickers}. We automatically generate a URL using the company code obtained. From the SEC website, either filings are downloaded as \texttt{TXT}, \texttt{HTML}, or \texttt{XBRL} using the generated URL. At this stage, approximately 6.5\% (or 543) data points corresponding to approximately 9.4\% (or 17) documents were dropped, either due to lack of mapping or non-availability of older documents. Further, the conversion of the downloaded files to PDF caused a loss of 117 data points (19 unique documents) due to formatting issues.

\begin{table}[t]
\footnotesize
% \small
\begin{tabular}{|p{7cm}|}
    \hline
    \\
    \textbf{ID:} \texttt{C/2017/page\_328.pdf-1} \\
    
    \\
    \mtcb{1}{\bf Context:}\\

        \\Performance graph comparison of five-year cumulative total return the following graph and table compare the cumulative total return on Citi 2019s common stock, which is listed on the NYSE under the ticker symbol 201cc 201d and held by 65691 common stockholders of record as of January 31, 2018, with the cumulative total return of the S\&P 500 index and the S\&P financial index over the five-year period through December 31, 2017. The graph and table assume that \$ 100 was invested on December 31, 2012 in Citi 2019s common stock, the S\&P 500 index and the S\&P financial index, and that all dividends were reinvested . comparison of five-year cumulative total return for the years ended date Citi S\&P 500 financials.\\
    
        \\| DATE | CITI | S\&P 500 | S\&P FINANCIALS |\\
        | :--- | :--- | :--- | :--- |\\
        | 31-Dec-2012 | 100.0 | 100.0 | 100.0 |\\
        | 31-Dec-2013 | 131.8 | 132.4 | 135.6 |\\
        | 31-Dec-2014 | 137.0 | 150.5 | 156.2 |\\
        | 31-Dec-2015 | 131.4 | 152.6 | 153.9 |\\
        | 31-Dec-2016 | 152.3 | 170.8 | 188.9 |\\
        | 31-Dec-2017 | 193.5 | 208.1 | 230.9 |\\
        \\
\mtcb{1}{\bf Question: }\\
\\
\it What was the percentage cumulative total return for the five year period ended 31-dec-2017 of citi common stock?\\
\\
\mtcb{1}{\bf Answer:}\\
\\
\multicolumn{1}{|c|}{93.5\%}\\
    \\
\hline
\end{tabular}

\caption{Example from {\bf FinQA} dataset. The context provided here has been formatted from the original dataset values.}
\label{tab:appendix_example_finqa}
\end{table}

\begin{table*}
\footnotesize
% \small

\begin{tabular}{|p{\linewidth}|}
    \hline
    \\
    \mtcb{1}{\bf Context:}\\
    
    Table of Contents \\
    UNITED STATES SECURITIES AND EXCHANGE COMMISSION \\
    \# ANNUAL REPORT PURSUANT TO SECTION 13 OR 15(d) OF THE SECURITIES EXCHANGE ACT of 1934 \\
    For the Fiscal Year Ended December 30, 2006 \\
    Commission file number 1-4171 \\
    \# Kellogg Company\\
    (Exact Name of Registrant as Specified in its Charter)\\
    Delaware (State of Incorporation) (I.R.S. Employer Identification No.) One Kellogg Square (Address of Principal Executive Offices) Securities registered pursuant to Section 12(b) of the Securities Act: Title of each class: Name of each exchange on which registered: \\
    $\cdots$ \\
    The Consolidated Financial Statements and related Notes, together with Management's Report on Internal Control over Financial Reporting, and the Report thereon of Pricewaterhouse Coopers LLP dated February 23, 2007, are included herein in Part II, Item 8. \\
    \# (a) 1. Consolidated Financial Statements Consolidated Statement of Earnings for the years ended December 30, 2006, December 31, 2005 and January 1, 2005. Consolidated Statement of Shareholders' Equity for the years ended December 30, 2006, December 31, 2005 and January 1, 2005. Notes to Consolidated Financial Statements. \\
    \# (a) 2. Consolidated Financial Statement Schedule All financial statement schedules are omitted because they are not applicable or the required information is shown in the financial statements or the notes thereto. \\
    \# (a) 3. Exhibits required to be filed by Item 601 of Regulation S-K The information called for by this Item is incorporated herein by reference from the Exhibit Index on pages 61 through 64 of this Report. Pursuant to the requirements of Section 13 or 15(d) of the Securities Exchange Act of 1934, the Registrant has duly caused this Report to be signed on its behalf by the undersigned, thereunto duly authorized, this 23rd day of February, 2007. Pursuant to the requirements of the Securities Exchange Act of 1934, this Report has been signed below by the following persons on behalf of the Registrant and in the capacities and on the dates indicated. Electronic(E), | 10.48 | | IBRF |\\
    | :--- | :--- | :--- | \\
    | | Commission file number 1-4171.* | |\\
    |  21.01 | Domestic and Foreign Subsidiaries of Kellogg. | E |\\
    |  23.01 | Consent of Independent Registered Public Accounting Firm. | E | \\
    |  24.01 | Powers of Attorney authorizing Gary H. Pilnick to execute our Annual Report on Form 10-K for the fiscal year ended December 30, 2006, on behalf of the Board of Directors, and each of them. | E | \\
    |  31.1 | Rule 13a-14(a)/15d-14(a) Certification by A.D. David Mackay. | E |\\
    |  31.2 | Rule 13a-14(a)/15d-14(a) Certification by John A. Bryant. | E |\\
    |  32.1 | Section 1350 Certification by A.D. David Mackay. | E | \\
    |  32.2 |  Section 1350 Certification by John A. Bryant. | E | \\
    % * A management contract or compensatory plan required to be filed with this Report. We agree to furnish to the Securities and Exchange Commission, upon its request, a copy of any instrument defining the rights of holders of long-term debt of Kellogg and our subsidiaries and any of our unconsolidated subsidiaries for which Financial Statements are required to be filed. We will furnish any of our shareowners a copy of any of the above Exhibits not included herein upon the written request of such shareowner and the payment to Kellogg of the reasonable expenses incurred in furnishing such copy or copies. \\

\\\mtcb{1}{\bf Question: }\\
\\
\multicolumn{1}{|c|}{\it What was the average cash flow from 2004 to 2006?}\\ 
\\
\mtcb{1}{\bf Program:}\\
\begin{lstlisting}[backgroundcolor=\color{white},numbersep=5pt,xleftmargin = 2em]
    net_cash_2006 = 957.4
    net_cash_2005 = 769.1
    net_cash_2004 = 950.4
    total_net_cash = net_cash_2006 + net_cash_2005 + net_cash_2004
    average_net_cash = total_net_cash / 3
    answer = average_net_cash
\end{lstlisting}
% \lstset{firstnumber=19}\hline
\\
\mtcb{1}{\bf Answer: } \\
\\
\multicolumn{1}{|c|}{892.3}\\
\hline
\end{tabular}

\caption{Examples from {\bf DocFinQA} dataset with text and tables from entire SEC document as context (truncated for legibility), question, associated program and answer. A full report can be founded here \url{https://www.annualreports.com/HostedData/AnnualReportArchive/k/NYSE_K_2006.pdf}}
\label{tab:appendix_example_docfinqa}
\end{table*}

\clearpage

\section{Parsing SEC Filings}
\label{app:sec-parsing}

Since each filing contains many tables, maintaining the structure and order during extraction is critical for numerical reasoning. We convert each HTML-formatted filing to PDF format and use a finance-specific PDF extractor 
%\href{https://kensho.com/extract}{Kensho Extract} 
to parse the filing into markdown format. 
This process ensures that: (i) our dataset is grounded in the relevant financial documentation and (ii) all the tables in the filings are parsed with high precision into a consistent format without any HTML-tag noise.

\begin{figure}[!h]
    \centering
    \includegraphics[width=0.85\linewidth]{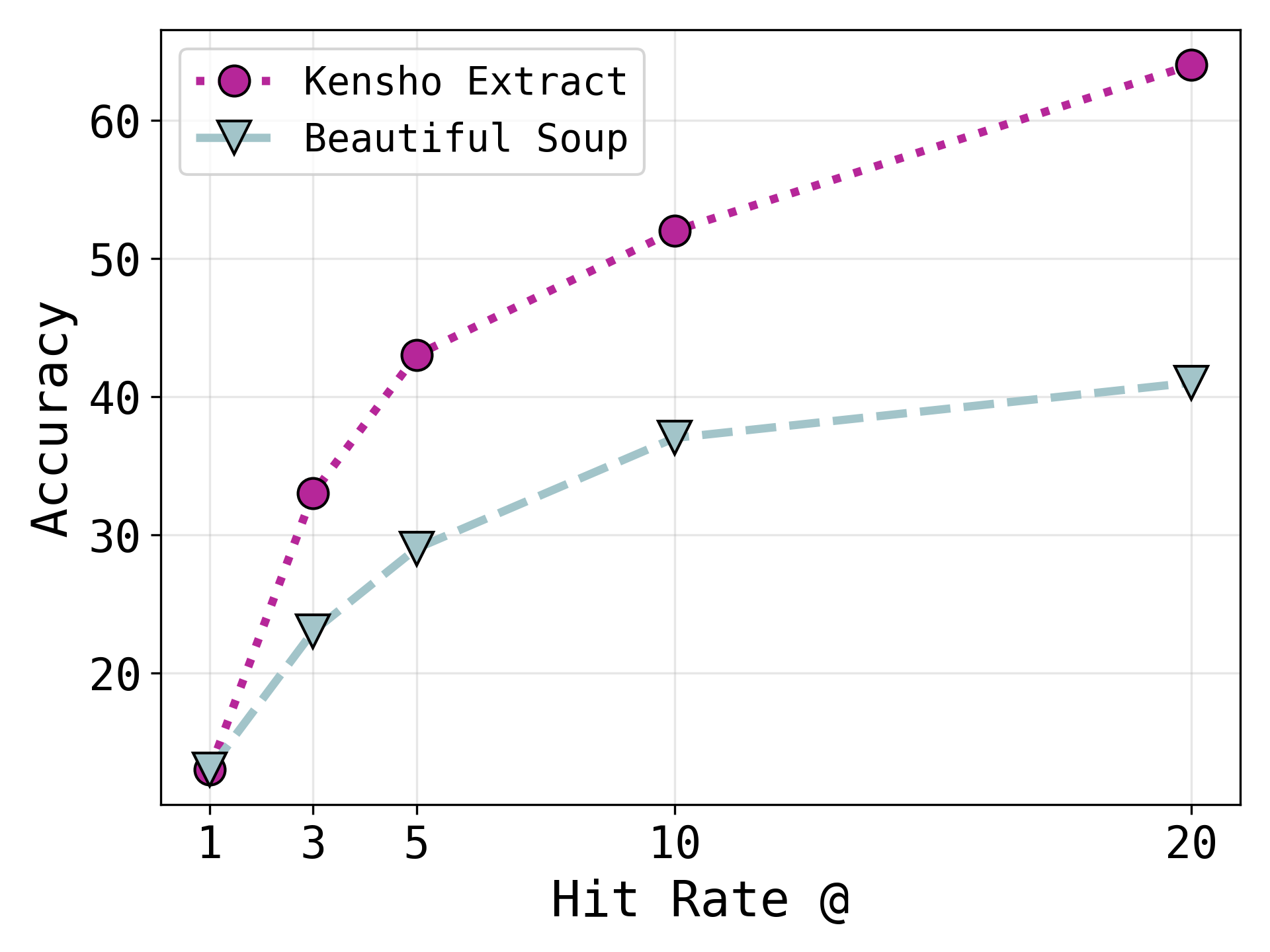}
    \caption{Accuracy for varying HR@ for two context extraction methods.}
    \label{fig:kensho-beautiful-soup}
\end{figure}

We explore different methods for parsing SEC filings consisting of HTML and XML markup into text and markdown tables for use in our QA systems. To evaluate parsing strategies, we measure HR@k (Hit Rate @ k) when searching for the gold chunk among all document chunks for a single document using the FinQA question as the search query. Queries and document chunks are encoded with OpenAI's ADA model. We compare BeautifulSoup, a standard library for manipulating HTML and XML formatted data, and Kensho Extract, a finance-specific text and table extraction model.\footnote{Passing the raw HTML/XML to the language model produces near-zero performance.} Figure~\ref{fig:kensho-beautiful-soup} shows the performance of these two methods. 

Additionally, we note a better downstream performance of finance-specific models with Kensho Extract retrieved-context compared to that of Beautiful Soup. Qualitative analysis of the different parsers reveals that Kensho Extract is better at structuring the tables used in financial documents, resulting in better readability which seems to extend to the encodings.

\newpage

\section{Code Conversion}
\label{app:code-generation}

Figure \ref{fig:code-generation} shows the steps of converting  (a) derivation of the result in FinQA into (b) dummy Python code with dummy variable names, and finally transforming it to (c) a meaningful Python program in DocFinQA following the work by \citet{bizbench}.

\begin{figure}[!h]
\begin{minipage}{\linewidth}
    \centering  
    (a) \\
    $subtract(34.8, 1.2), divide(\#0, 34.8)$ \\
    \vspace{1em}
    (b) \\
    $a = 34.8 - 1.2$\\
    $b = a / 34.8$\\
    $c = b * 100$ \\
    \vspace{1em}
    (c) \\
    $payments\_decrease = 34.8 - 1.2$\\
    $change = payments\_decrease / 34.8$\\
    $answer = change * 100$ 
\end{minipage}

\caption{Example of code conversion. (a) Original FinQA's derivation. (b) Dummy Python Program (c) Meaningful Python Code in DocFinQA.}
\label{fig:code-generation}
\end{figure}    

\section{Model Details}
\label{app:model-detail}

In this work, we used the base models of Falcon, MPT, Llama 2, CodeLlama, and Mistral throughout our work. These models were not trained with supervised finetuning or reinforcement learning human feedback. The GPT-3.5 model employed in this study is \texttt{gpt-3.5-turbo-0613} while the GPT-4 model used is \texttt{gpt-4-1106-preview}.

We also included the Llama 2/7B + SFT that was finetuned on the training set of DocFinQA with golden chunk from FinQA ($c^{golden}$). The finetuning process takes 3 epochs with a batch size of 32. We use the context provided by the FinQA dataset as the input due to the limited maximum token length of the model. The maximum token length is set to 2048. The model is finetuned on 8 x Nvidia A100-80GB GPUs. We use AdamW optimizer with learning rate of 2e-6. The training process takes 4 hours to complete.

\newpage

\section{Distribution of question by question types}
\label{app:questio-dist}

\begin{figure}[!h]
    \includegraphics[width=\linewidth]{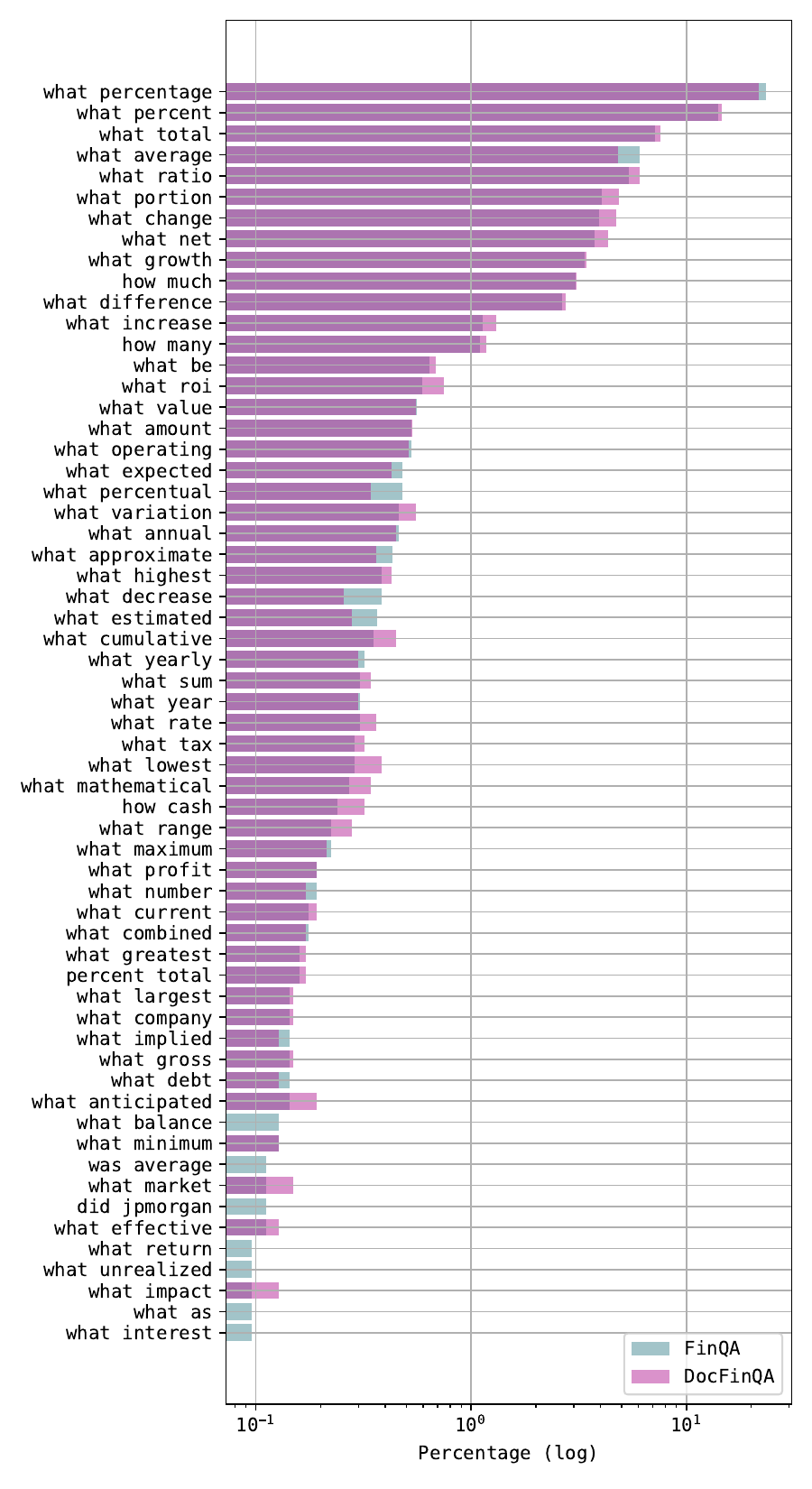}
    \caption{Distribution of questions grouped by question types in the original FinQA and DocFinQA. The x-axis (percentage) is presented in log scale to magnify the differences between the two sets.}
    \label{fig:question-dist}
\end{figure}

Figure \ref{fig:question-dist} shows the distribution by question types of the dataset before (FinQA, green) and after (DocFinQA, purple) the automatic data selection.

\newpage

\section{Performance of retrieval methods}
\label{app:bm25-included}

\begin{figure}[!h]
    \centering
    \includegraphics[width=.85\linewidth]{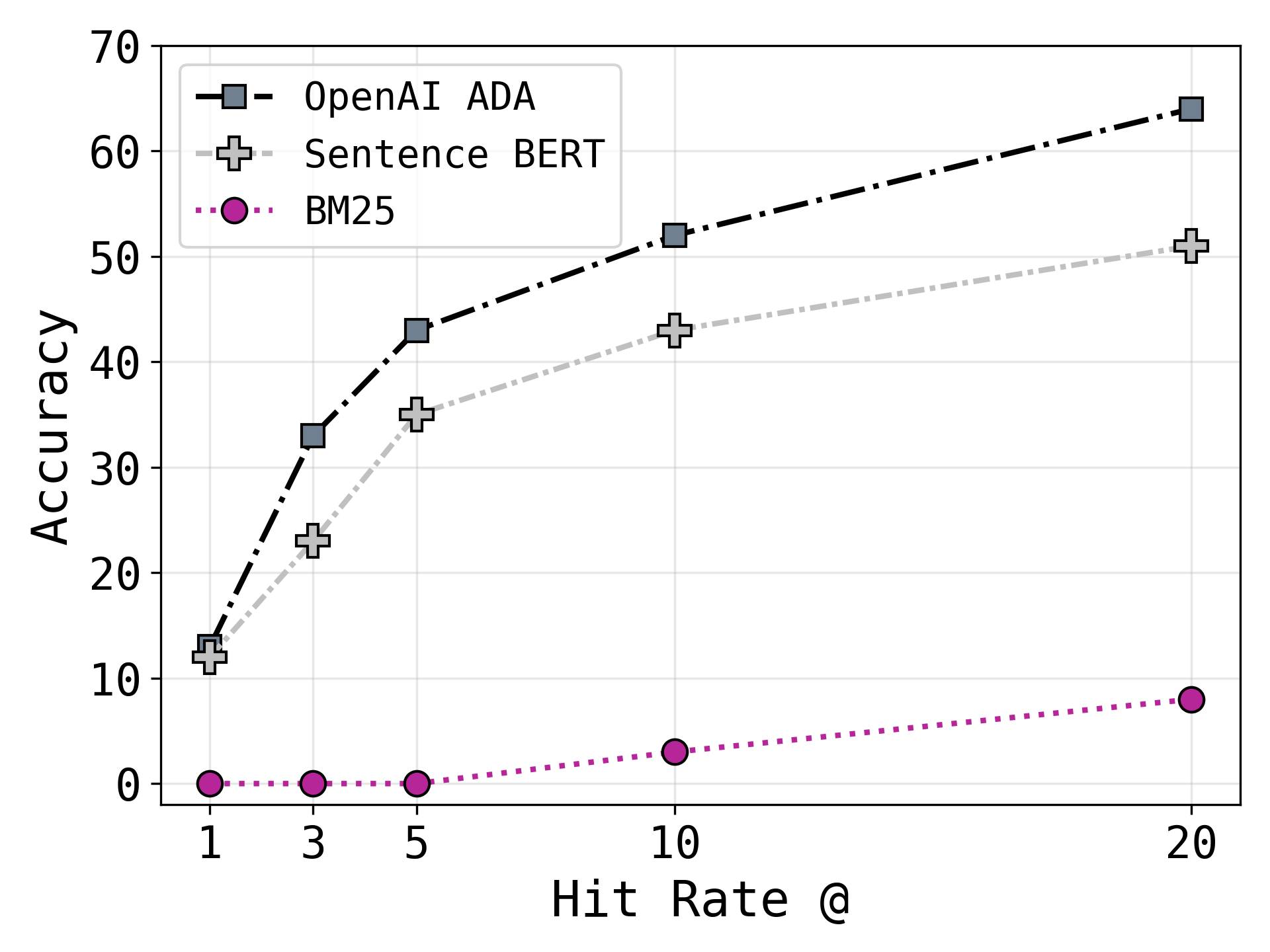}
    \caption{Accuracy for varying HR@ for three search methods on the development set}
    \label{fig:bm25-included}
\end{figure}

Figure \ref{fig:bm25-included} shows a pilot study comparing dense retrieval with OpenAI ADA and Sentence BERT versus sparse retrieval (BM 25) on the development set. We can see that the dense retrieval model offers a much higher hit ratio.

Figure \ref{fig:prompt-template} shows the prompt template with in-context learning that we used.

\begin{figure}[!h]
\boxed{
\begin{minipage}{0.9\linewidth}
\tt
\scriptsize
Context: \{golden chunk\}\\
Question: \{question\}\\
Python Program: \{program\}\\
Answer: \{answer\}\\

Context: \{golden chunk\}\\
Question: \{question\}\\
Python Program: \{program\}\\
Answer: \{answer\}\\

Context: \{golden chunk\}\\
Question: \{question\}\\
Python Program: \{program\}\\
Answer: \{answer\} \\

Context: \{first chunk\}\\\
\{second chunk\}\\
\{third chunk\}\\
Question: \{question\}\\
Python Program: 
\end{minipage}
}
\caption{Prompt template with Top-3 context and 3-shot In-Context Learning.}
\label{fig:prompt-template}
\end{figure}

\newpage

\section{ColBERT Finetuning}
\label{app:colber-finetuning}
\begin{figure}[!h]
    \centering
    \includegraphics[width=0.85\linewidth]{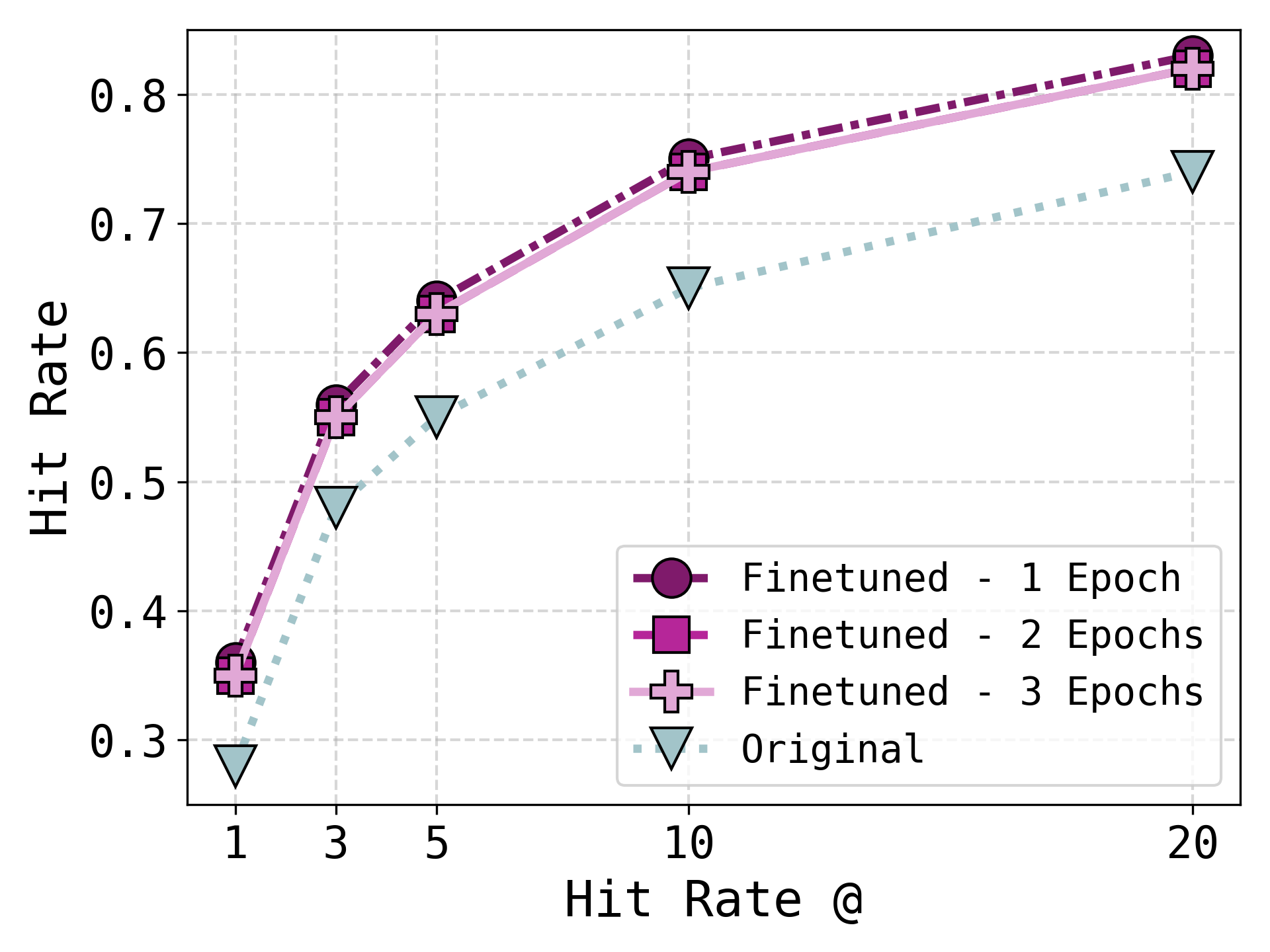}
    \caption{Hit rate of different ColBERT variants on the development set of DocFinQA.}
\label{fig:colbert-finetuning}
\end{figure}

We finetune the original ColBERT v1 model on the train set of DocFinQA. For each data point, we perform chunking and alignment to generate one golden chunk and $n-1$ negative chunks. For training, we generate a list of tuples (qid, pid+, pid-), where qid refers to the question, pid+ refers to the golden chunk and pid- refers to each of the negative chunks in that document. We train the model for a total of 3 epochs and store the checkpoints at the end of each epoch. The hit rate of the Finetuned ColBERT model after each epoch on the development set is shown in Figure~\ref{fig:colbert-finetuning}. We observe that after the first epoch, additional finetuning does not show any performance improvement. The Finetuned ColBERT model referred to in this study thus uses the weights after the first epoch of training.

\newpage

\section{Few-shot Settings}
\label{app:few-shot}

\begin{figure}[!h]
    \centering
    \includegraphics[width=\linewidth]{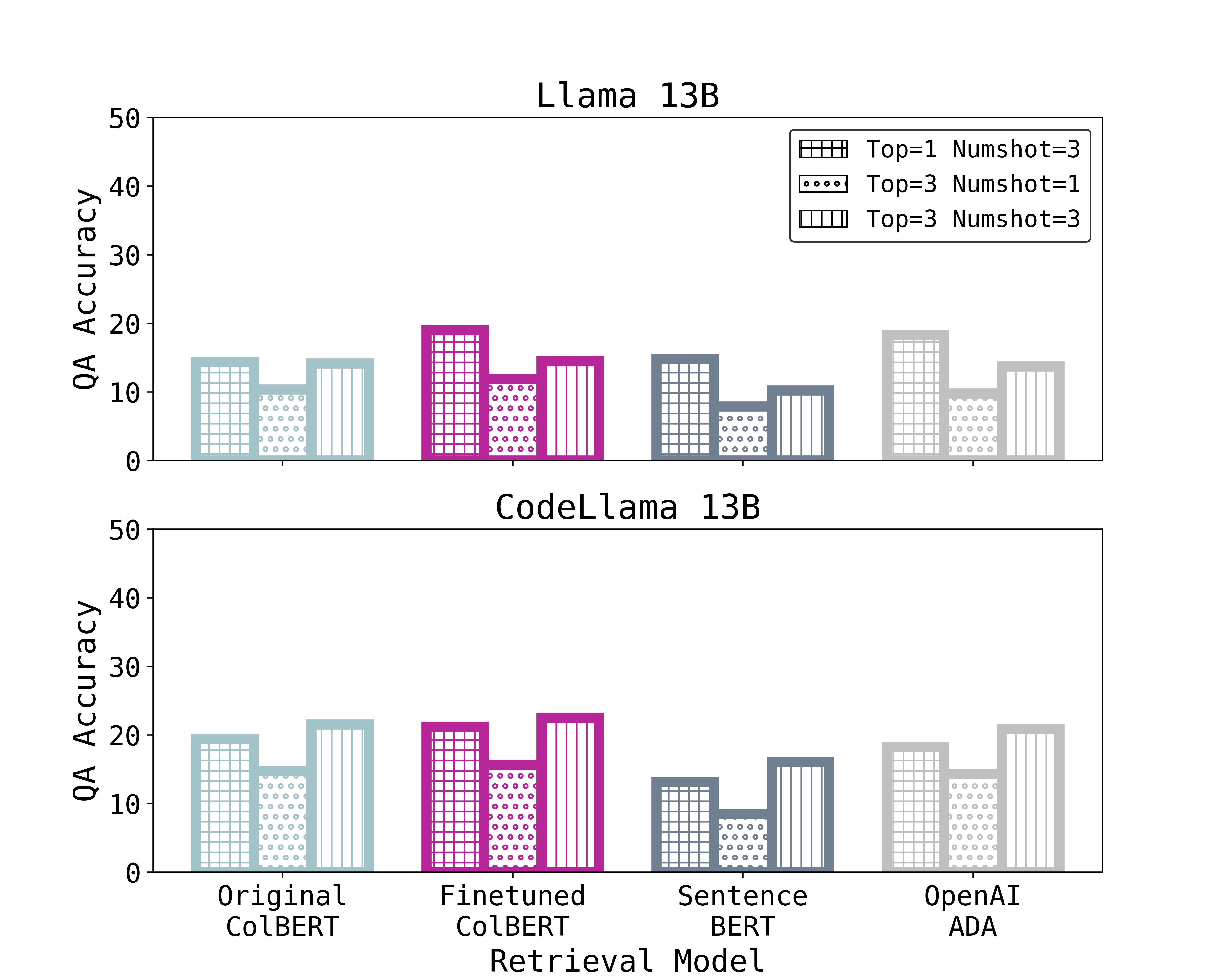}
    \caption{A QA performance plot on the development set of DocFinQA for the Llama 2 13B and CodeLlama 2 13B models for each of the 12 configurations}
    \label{fig:topk-numshot}
\end{figure}

Due to the limited context length of the LLMs, the number of few-shot demonstrations and the number of chunks fed into the In-Context Learning must be optimized. We explore 3 settings of the number of few-shot examples and 4 settings of the number of chunks used as context in the query. Figure \ref{fig:topk-numshot} shows the performance of these settings in retrieval and answered by LLama 13B and CodeLLaMa 13B on the development set. We see that a higher number of few-shot examples (numshot=3) yield consistently better performance compared to a lower one (numshot=1).

\newpage

\section{Golden Chunk Position}

\begin{figure}[!h]
    \centering
    \includegraphics[width=0.85\linewidth]{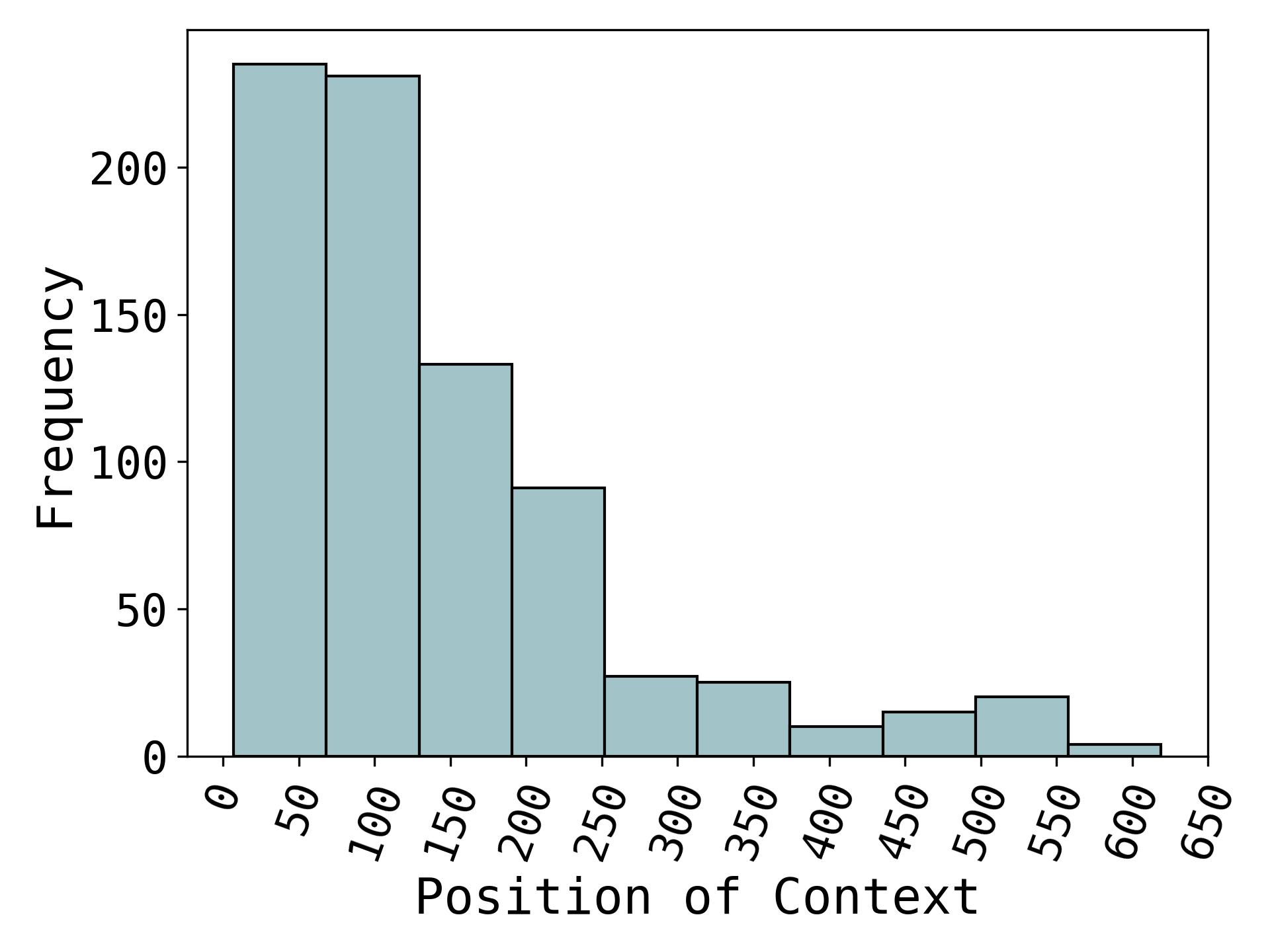}
    \caption{Histogram of the position of the FinQA context in the original SEC filing that is split into chunks of size 2750.}
    \label{fig:Chunk_Position}
\end{figure}

Figure~\ref{fig:Chunk_Position} shows the distribution of the position golden chunk with the documents. We see that most of the golden chunks appear within the first 250 chunks (approximately 125K tokens which can be fed into the newest generative models). Nonetheless, there are a substantial number of questions that the golden chunk appears beyond this threshold.

\section{Human Evaluation Setting}

\label{app:human-eval}

We recruited three data professionals with 4-5 years of experience working with financial documents, including but not limited to 10-K filings, to estimate human evaluation. The professionals were provided with the entire document in PDF format, maintaining the SEC's original format for ease of reading. They were allowed to use the keyword-search feature of PDF reader applications and a simple calculator for basic arithmetic operations required for this task. On average, the professionals spent 25 minutes per question.

\end{document}